  \documentclass{gretsi}
\usepackage{amsmath,graphicx}
\usepackage{algorithm}
\usepackage[noend]{algpseudocode}
\usepackage{amssymb,bm}
\usepackage{float}
\usepackage{colortbl}
\usepackage{subfigure}
\usepackage{amsthm}
\usepackage[dvipsnames]{xcolor}
\definecolor{mygreen}{RGB}{0, 199, 0}
\definecolor{myorange}{RGB}{250, 100, 0}
\definecolor{myred}{RGB}{200, 0, 0}
\definecolor{myblue}{RGB}{30, 144, 255}
\definecolor{mylightskyblue}{RGB}{135, 206, 250}
\definecolor{myskyblue}{RGB}{0, 191, 255}
\definecolor{mypowderblue}{RGB}{176, 196, 222}
\newtheorem{theorem}{Th\'eor\`eme}[section]
\newtheorem{proposition}[theorem]{Proposition}

\newtheorem*{preuve}{Preuve}

\usepackage[english,french]{babel}   
 
\usepackage[T1]{fontenc}

\titre{Algorithme EM r\'egularis\'e}

\auteur{\coord{Pierre}{HOUDOUIN}{1},
    \coord{Matthieu}{JONCKHEERE}{2}, \coord{Fr\'ed\'eric}{PASCAL}{1}, \coord{Esa}{OLLILA}{3}}

\adresse{\affil{1}{Universit\'e Paris-Saclay, CNRS, CentraleSup\'elec,
Laboratoire des signaux et syst\`emes,
91190, Gif-sur-Yvette, France}
         \affil{2}{LAAS-CNRS, Universit\'e de Toulouse,
31077 Toulouse, France}
        \affil{3}{Aalto University,
        Helsinki, Finland}}

\email{pierre.houdouin@centralesupelec.fr,
frederic.pascal@centralesupelec.fr,
mjonckheer@laas.fr,
esa.ollila@aalto.fi}

\resumefrancais{L'algorithme Expectation-Maximization (EM) est un algorithme it\'eratif, notamment utilis\'e pour estimer les maximum de vraisemblance de donn\'ees issues d'un mod\`ele de m\'elange gaussien (GMM). Lorsque la taille de l'\'echantillon des donn\'ees est proche de leur dimension, les estimations successives de la matrice  de covariance peuvent \^etre singuli\`eres ou mal conditionn\'ees, entraînant une baisse des performances. Nous pr\'esentons dans ce papier une nouvelle version r\'egularis\'ee de l'algorithme EM adapté au cas o\`u le rapport entre taille de l'\'echantillon et la dimension est faible. Cette m\'ethode maximise une version p\'enalis\'ee de la vraisemblance de l'algorithme EM-GMM qui assure que les covariances soient toujours d\'efinies positives. Des tests sur données réelles illustrent enfin l'intérêt de cette approche pour un problème de clustering.}

\resumeanglais{Expectation-Maximization (EM) algorithm is a widely used iterative algorithm for computing maximum likelihood estimate when dealing with Gaussian Mixture Model (GMM). When the sample size is smaller than the data dimension, this could lead to a singular or poorly conditioned covariance matrix and, thus, to performance reduction. This paper presents a regularized version of the EM algorithm that efficiently uses prior knowledge to cope with a small sample size. This method aims to maximize a penalized GMM likelihood where regularized estimation may ensure positive definiteness of covariance matrix updates by shrinking the estimators towards some structured target covariance matrices. Finally, experiments on real data highlight the good performance of the proposed algorithm for clustering purposes.}

\begin{document}
\maketitle

\section{Introduction}
\label{sec:intro}

L'algorithme EM \cite{Dempster2022Maximum} est un algorithme fr\'equemment utilis\'e en apprentissage non supervis\'e et en mod\'elisation statistique permettant de trouver un maximum local de la vrai\-semblance de donn\'ees non labellis\'ees et d'estimer les labels associ\'es. En proc\'edant it\'erativement, cet algorithme estime les param\`etres inconnus du mod\`ele qui augmentent l'esp\'erance de la vraisemblance des donn\'ees compl\'et\'ees grâce aux param\`etres de l'it\'eration pr\'ec\'edente. 

Historiquement \'elabor\'e pour les mod\`eles de m\'elange de gaussien (GMM) \cite{Guorong2001EM}, l'algorithme a \'et\'e \'etendu aux distributions de Student par \cite{Ingrassia2012Studies} pour mieux faire face aux donn\'ees aberrantes et aux donn\'ees \`a queues lourdes. Plus r\'ecemment, une g\'en\'eralisation aux distributions elliptiques sym\'etriques a \'et\'e d\'evelopp\'ee (\cite{roizman2019flexible} pour le clustering et \cite{houdouin2022robust} pour la classification). 

En traitement du signal, la dimension $m$ des donn\'ees est souvent \'elev\'ee par rapport à leur nombre $n$ : $n \sim m$. Dans de telles conditions, des probl\`emes de convergence surviennent lors de l'estimation des matrices de covariance qui ne sont plus forc\'ement bien conditionn\'ees ou m\^eme inversibles \`a chaque it\'eration. 
L'estimation de matrices de covariance r\'egularis\'ees est une technique couramment utilis\'ee pour surmonter cette difficult\'e dans les mod\`eles de clustering \cite{Ying2014Regularized,pascal2014generalized,yi2020shrinking}.

En 2022, \cite{houdouin2022regularized} pr\'esente une nouvelle version r\'egularis\'ee de l'algorithme EM, RG-EM, qui utilise une nouvelle p\'enalisation de la vraisemblance permettant de tirer parti de la structure sous-jacente suppos\'ee des matrices de covariance. \cite{houdouin2022regularized} montre que des estimateurs mieux conditionn\'es sont obte\-nus, avec de meilleures performances de clustering dans les r\'egimes o\`u la dimension est elev\'ee par rapport au nombre de donn\'ees. Nous proposons ici d'\'evaluer les performances de RG-EM sur des donn\'ees r\'eelles. La structure du papier est la suivante : la section \ref{sec:2} rappelle les \'el\'ements th\'eoriques de l'algorithme, la section \ref{sec:3} contient les exp\'eriences sur donn\'ees r\'eelles et les conclusions, remarques et perspectives sont \'etablies dans la section \ref{sec:conclusions}.

\section{Algorithme EM r\'egularis\'e}
\label{sec:2}

On suppose que chaque observation $\mathbf{x}_i \in \mathbb{R}^m$ est issue d'un GMM o\`u chaque cluster $\mathcal{C}_k$ a son propre vecteur moyenne $\boldsymbol{\mu}_k \in \mathbb{R}^m$, sa propre matrice de covariance sym\'etrique d\'efinie positive $\boldsymbol{\Sigma}_k \in \mathbb{R}^{m \times m}$ et sa probabilit\'e d'appartenance $\pi_k \in [0, 1]$ avec $\sum_k \pi_k=1$. La densit\'e de probabilit\'e de $\mathbf{x}_i$ s'\'ecrit alors : 

$$
f(\mathbf{x}_i |\boldsymbol{\theta} ) =  (2\pi)^{-\frac{m}{2}} \sum_{k=1}^K \pi_k \left|\mathbf{\Sigma}_k \right|^{-\frac{1}{2}}  e^{ -\frac{1}{2} (\mathbf{x}_i-\boldsymbol{\mu}_k)^\top \mathbf{\Sigma}_k^{-1} (\mathbf{x}_i-\boldsymbol{\mu}_k)}
$$
avec $\boldsymbol{\theta} = (\pi_1,...,\pi_K,\boldsymbol{\mu}_1,...,\boldsymbol{\mu}_K,\mathbf{\Sigma}_1,...,\mathbf{\Sigma}_K)$, le vecteur de tous les param\`etres inconnus.

Supposons égalemet qu'une informatio a priori sur la structure des matrices de covariance de chaque cluster est disponible : e.g., elles sont proches de matrices cibles $\mathbf{T}_k$, $k=1,\ldots,K$. On exploite cette structure en p\'enalisant la vraisemblance avec la divergence de Kullback-Leibler (d\'efinie dans \cite{Ying2014Regularized}) entre chaque $\mathbf{\Sigma}_k$ et $\mathbf{T}_k$ :
\[
\Pi_{\textup{KL}}(\boldsymbol{\Sigma}_k,\mathbf{T}_k) =  \frac 1 2 \big(\mathrm{tr}(\boldsymbol{\Sigma}_k^{-1} \mathbf{T}_k)  - \log | \boldsymbol{\Sigma}_k^{-1} \mathbf{T}_k | - m \big).
\]

Soit $\mathbf{X} = (\mathbf{x}_1 \ \ \cdots \mathbf{x}_n)$ la matrice des donn\'ees issues de notre mod\`ele GMM, notre vraisemblance p\'enalis\'ee est alors :

$$ 
\ell_{\boldsymbol{\eta}}( \boldsymbol{\theta} \vert \mathbf{X} )=  \ell(\mathbf{X}|\boldsymbol{\theta}) - \sum_{k=1}^K \eta_k \Pi_{\textup{KL}}(\mathbf{\Sigma}_k,\mathbf{T}_k)
$$

o\`u $\eta_1,...,\eta_K \geq 0 $ sont des param\`etres automatiquement ajust\'es.

\begin{proposition}
L'\'etape E de l'algorithme EM r\'egularis\'e n'est pas modifi\'ee, on a $\forall i \in  [\![1,n]\!], \forall k \in  [\![1,K]\!]$ :
\begin{equation}
p_{ik}^{(t)} = \frac{\hat{\pi}_k^{(t)} |\mathbf{\hat{\Sigma}}^{(t)}_k|^{-\frac{1}{2}} e^{-\frac{1}{2} (\mathbf{x}_i-\boldsymbol{\hat{\mu}}_k^{(t)})^\top \mathbf{\hat{\Sigma}}^{t-1}_k (\mathbf{x}_i-\boldsymbol{\hat{\mu}}_k^{(t)})}}{\sum_{j=1}^K \hat{\pi}_j^{(t)} |\mathbf{\hat{\Sigma}}^{t}_j|^{-\frac{1}{2}} e^{-\frac{1}{2} (\mathbf{x}_i-\boldsymbol{\hat{\mu}}_j^{(t)})^\top \mathbf{\hat{\Sigma}}^{t-1}_j (\mathbf{x}_i-\boldsymbol{\hat{\mu}}_j^{(t)})}} 
\end{equation}
\end{proposition}

\begin{preuve} 
 Voir \cite{houdouin2022regularized}   
\end{preuve}

\begin{proposition}
Les mises \`a jours de l'\'etape M sont les suivantes :
\begin{align*}
\pi_k^{(t+1)} &= \frac{1}{n} \sum_{i=1}^n p_{ik}^{(t)}\,, &
\boldsymbol{\hat{\mu}}_k^{(t+1)} 
= \sum_{i=1}^n w_{ik}^{(t)} \mathbf{x}_i
\end{align*}

\begin{equation*}
\hat{\boldsymbol{\Sigma}}_k^{(t+1)} = \beta_k^{(t+1)} \sum_{i=1}^n w_{ik}^{(t)}  (\mathbf{x}_i-\boldsymbol{\hat{\mu}}_k^{(t)})( \mathbf{x}_i-\hat{\boldsymbol{\mu}}_k^{(t)})^\top + (1-\beta_k^{(t+1)})\mathbf{T}_k, 
\end{equation*}
o\`u
$\displaystyle \beta_k^{(t+1)} = \frac{n\pi_k^{(t+1)}}{\eta_k + n\pi_k^{(t+1)}}$ et $\displaystyle w_{ik}^{(t)} = \frac{p_{ik}^{(t)}}{\sum_{i=1}^n p_{ik}^{(t)}}
$
\end{proposition}
\begin{preuve}
 Voir \cite{houdouin2022regularized}
\end{preuve}

\begin{algorithm}[!t]
\caption{$L$-fold validation crois\'ee de $\eta_k$ pour le cluster $k$} \label{alg:CV2}

\textbf{Entr\'ees :} L'\'echelle $\hat \theta_k^0$ et les indices $\mathcal{D}^0$ des \'el\'ements du cluster $k$. Un ensemble de candidats $\{ \eta_j \}_{j=1}^J$  pour la valeur du param\`etre de p\'enalisation.
\begin{algorithmic}[1]

\State S\'eparer $\mathcal D^0$ en $L$ sous-ensembles distints $\mathcal{D}_1,\ldots,\mathcal{D}_L$ t.q $\mathcal D^0 = \cup_{l=1}^L \mathcal D_l$ et initialiser $\mathbf{T}_k^0 = \hat \theta_k^0 \cdot \mathbf{I}_m$ comme matrice cible.
\State Initialiser $\mathrm{Err}_j \equiv \mathrm{Err}(\eta_j)=0$ pour $j=1,\ldots,J$.

    \For{$l \in [\![1,L]\!]$}
        \State Initialiser $ \mathcal D_{\textup{val}} = \mathcal{D}_l$ and $\mathcal D_{\textup{tr}} = \mathcal{D} /\ \mathcal{D}_{\textup{val}}$
         \State $\mathbf{S}_{\textup{val}} = \frac 1{|\mathcal{D}_{\textup{val}} |} \sum_{i \in \mathcal D_{\textup{val}} }  (\mathbf{x}_i-\bar{\mathbf{x}}_{\textup{val}})(\mathbf{x}_i- \bar{\mathbf{x}}_{\textup{val}})^\top$
        \State $\boldsymbol{\hat{\Sigma}}= \frac 1{|\mathcal{D}_{\textup{tr}} |} \sum_{i \in  \mathcal{D}_{\textup{tr}}} (\mathbf{x}_i-\bar{\mathbf{x}}_{\textup{tr}})(\mathbf{x}_i-\bar{\mathbf{x}}_{\textup{tr}})^\top$
       \For{$\eta \in \{\eta_1,\ldots,\eta_J\}$}
            \State $\hat{\boldsymbol{\Sigma}}_{\eta}= \frac{|\mathcal D_{\textup{tr}}|}{\eta+|\mathcal D_{\textup{tr}}|} \boldsymbol{\hat{\Sigma}} + \frac{\eta}{\eta+|\mathcal D_{\textup{tr}}|} \mathbf{T}_k^0$
            \State $\mathrm{Err}_{l} = \mathrm{Err}_{l} + \mathrm{tr}(\hat{\boldsymbol{\Sigma}}_{\eta}^{-1}\mathbf{S}_{\textup{val}}) + \log | \boldsymbol{\Sigma}_{\eta} | $
        \EndFor
 \EndFor
\State Choisir le $\eta_j$ qui minimise $ \{\mathrm{Err}(\eta_j)\}_{j=1}^J$ 

\end{algorithmic}
\end{algorithm}

La matrice cible $\mathbf{T}_k$ permet ainsi d'injecter des connaissances a priori sur $\mathbf{\Sigma}_k$ dans l'estimation. Si aucune information a priori n'est disponible, on peut choisir $\mathbf{T}_k = \hat \theta_k^0 \mathbf{I}_m$, ce qui permet simplement d'assurer le bon conditionnement des estimateurs. On utilise alors l'estimateur classique du param\`etre d'\'echelle $\theta_k = \mathrm{tr}(\boldsymbol{\Sigma}_k)/m$. Dans nos exp\'eriences, on utilise $\hat \theta_k^0 = \mathrm{tr}(\hat{\boldsymbol{\Sigma}}_k^0)/m$ o\`u $\hat{\boldsymbol{\Sigma}}_k^0$ est la valeur initiale de l'estimation de la matrice de covariance, obtenue gr\^ace \`a un premier clustering avec l'algorithme K-means. Dans l'algorithme EM, la valeur du param\`etre d'\'echelle est périodiquement mise \`a jour avec la 
nouvelle valeur de $\hat{\boldsymbol{\Sigma}}_k$.

Le choix du param\`etre de r\'egularisation est \'egalement essentiel. On utilise une s\'election par validation crois\'ee qui maximise la log-vraisemblance gaussienne \cite{yi2020shrinking}. Chaque $\eta_k$ est estim\'e ind\'ependamment parmi un ensemble de candidats $\{\eta_1,\ldots,\eta_J\}$ par la proc\'edure d\'ecrite dans l'algorithme \ref{alg:CV2}. 

\section{Exp\'eriences sur donn\'ees simul\'ees}
\label{sec:3}

L'algorithme EM r\'egularis\'e, RG-EM, est compar\'e \`a l'EM classique, noté G-EM, ainsi qu'\`a l'algorithme K-means. Les deux versions de l'EM sont impl\'ement\'ees et la version de Scikit-learn pour le K-means est utilisée. Afin que l'EM classique converge m\^eme lorsque la dimension est \'elev\'ee et qu'il y a peu de donn\'ees, on ajoute une r\'egularisation classique avec la matrice $\epsilon \,\mathbf{I}_m$ \`a chaque it\'eration. On utilise pour \textbf{K-means} $n_{init}=10$ et $max_{iter}=200$, pour \textbf{G-EM} $\epsilon=10^{-4}$ et $max_{iter}=40$ et pour \textbf{RG-EM} $L=5$ (Algorithme \ref{alg:CV2}) et $max_{iter}=40$. Comme indiqué dans la section 2, on utilise pour matrices cibles $\mathbf{T}_k = \mathrm{tr}(\hat{\boldsymbol{\Sigma}}_k^0)/m \mathbf{I}_m$. Pour \textbf{RG-EM}, on recalcule les $\eta_k$ optimaux toute les $10$ it\'erations. Les donn\'ees g\'en\'er\'ees \`a partir de distributions gaussiennes sont r\'eparties en $K=3$ clusters avec les priors $\pi_k=\frac{1}{3}$. Le vecteur moyenne est tir\'e al\'eatoirement sur la sph\`ere centr\'ee de rayon 2 tandis qu'on utilise une structure autor\'egressive pour les covariances. On choisit $\left(\boldsymbol{\Sigma}_k\right)_{i,j}= \rho_k^{|i-j|}$ avec les coefficients $0.8$, $0.5$ et $0.2$. Cela traduit une structure autor\'egressive dans les donn\'ees. 

\begin{figure}[!ht]
\centering
\subfigure[n = 1000\label{fig:1a}]{\includegraphics[scale=0.13]{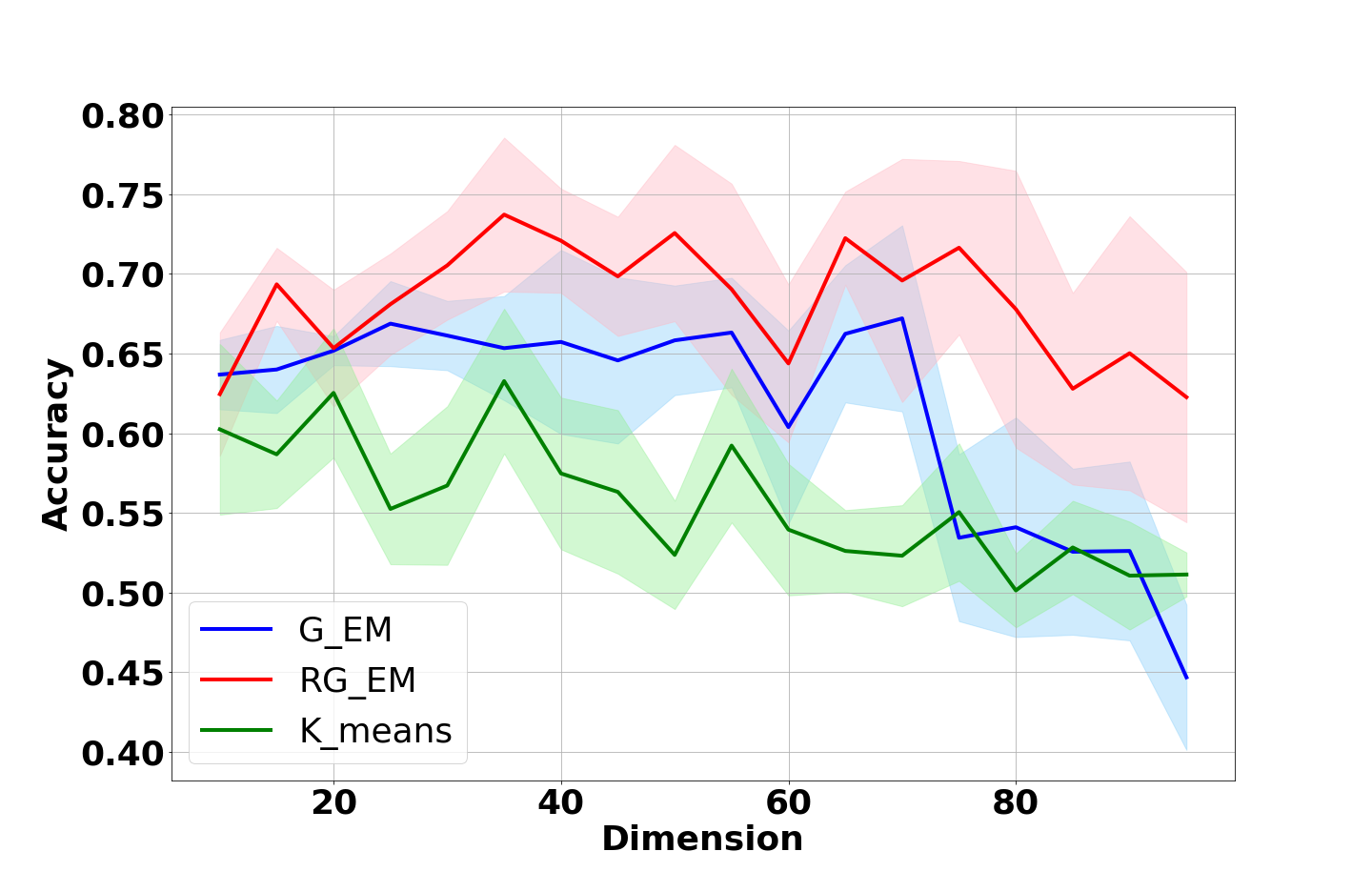}}
\subfigure[n = 500\label{fig:1b}]{\includegraphics[scale=0.13]{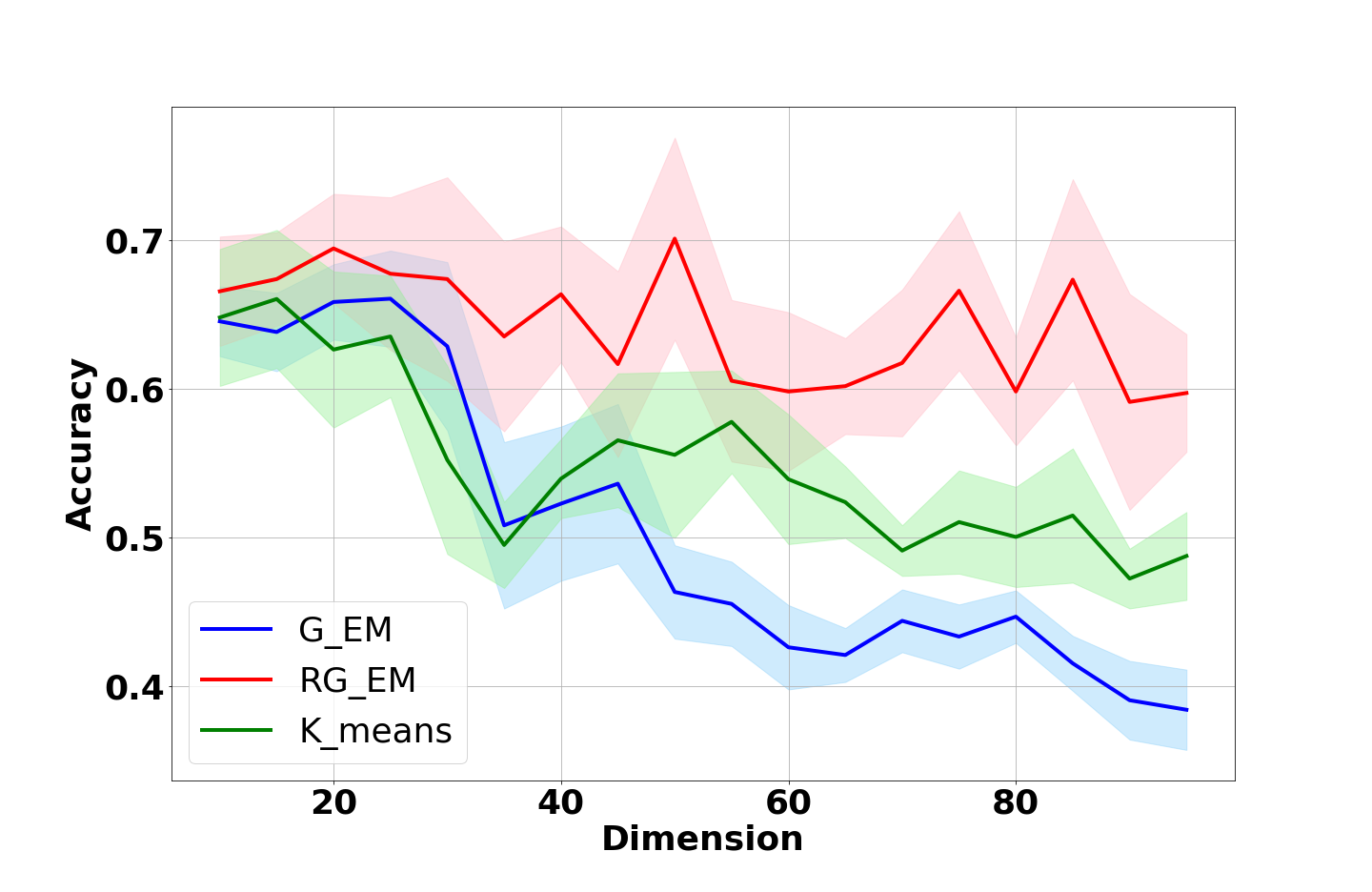}}
\caption{Evolution de la pr\'ecision en fonction de la dimension}
\label{fig:1} 
\end{figure}

On teste deux configurations avec, respectivement, $n=1000$ et $n=500$. On \'evalue la performance des mod\`eles en calculant leur pr\'ecision. Pour calculer la pr\'ecision en clustering, on commence par calculer la matrice de confusion, puis on permute les colonnes de sorte \`a maximiser la somme des \'el\'ements diagonaux. Les r\'esultats sont pr\'esent\'es en figure \ref{fig:1}.

Dans les deux configurations, il y a une dimension \`a partir de laquelle les performances de l'EM classique chutent et cela correspond au ratio $\frac{n}{m} \approx 14$. A l'inverse, l'EM r\'egularis\'e parvient \`a conserver des performances similaires entre la dimension $10$ et la dimension $100$. En effet, les matrices de covariance des clusters ont une structure proche d'une identit\'e, surtout lorsque $\rho$ est proche de 0. La matrice cible choisie s'av\`ere donc ici particuli\`erement pertinent
\section{Exp\'eriences sur donn\'ees r\'eelles}
\label{sec:4}

On teste chaque m\'ethode sur des jeux de donn\'ees r\'eelles issus de l'UCI machine learning repository \cite{UCI}. Deux datasets sont utilis\'es : 
\begin{itemize}
    \item \textbf{Ionosphere} : $n=351$, $p=34$ et $K=2$
    \item \textbf{Breast cancer} : $n=699$, $p=9$ et $K=2$
\end{itemize}

\begin{figure}[!ht]
\centering
\subfigure[ionosphere\label{fig:2a}]{\includegraphics[scale=0.13]{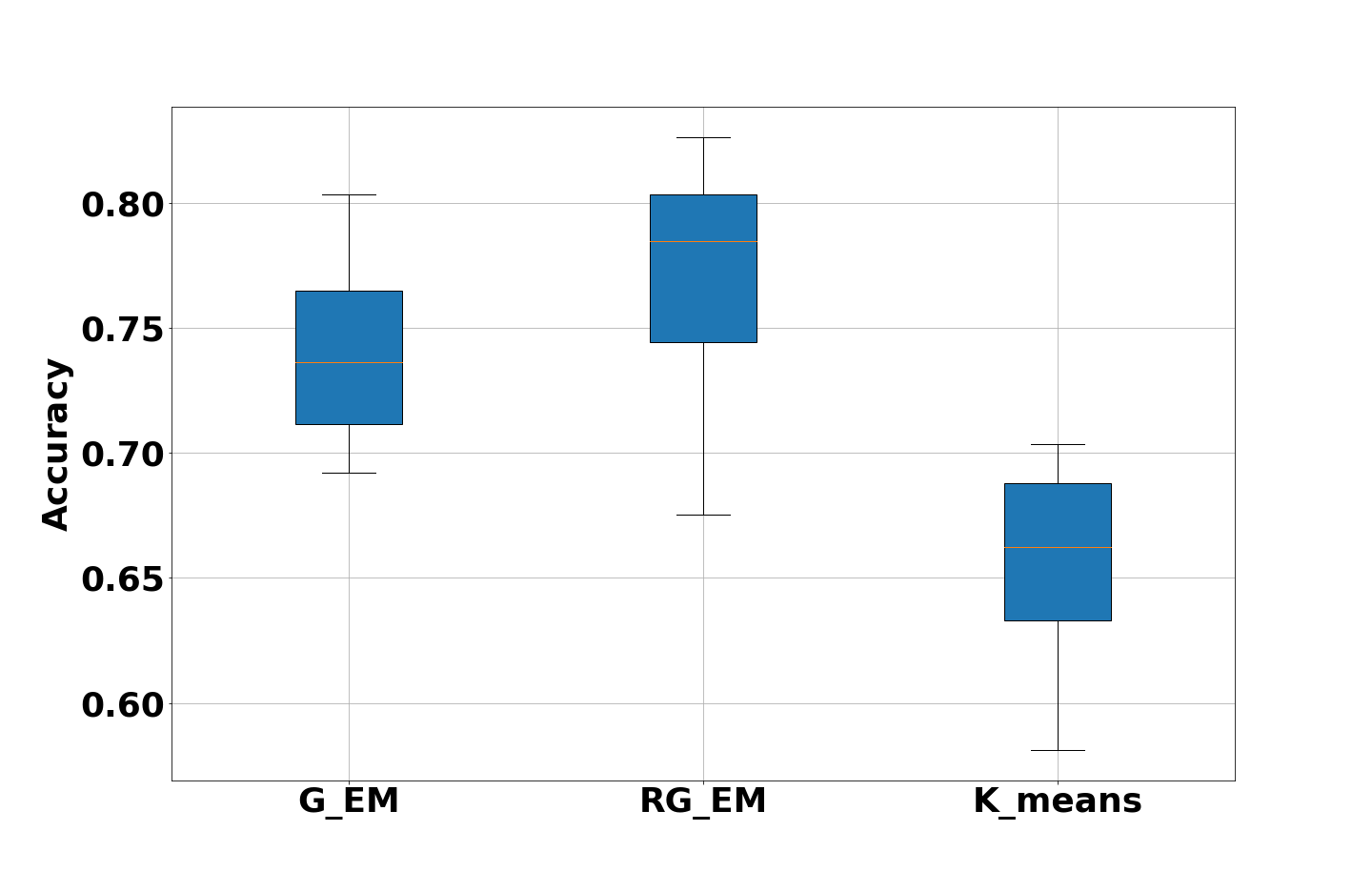}}
\subfigure[Breast cancer\label{fig:2b}]{\includegraphics[scale=0.13]{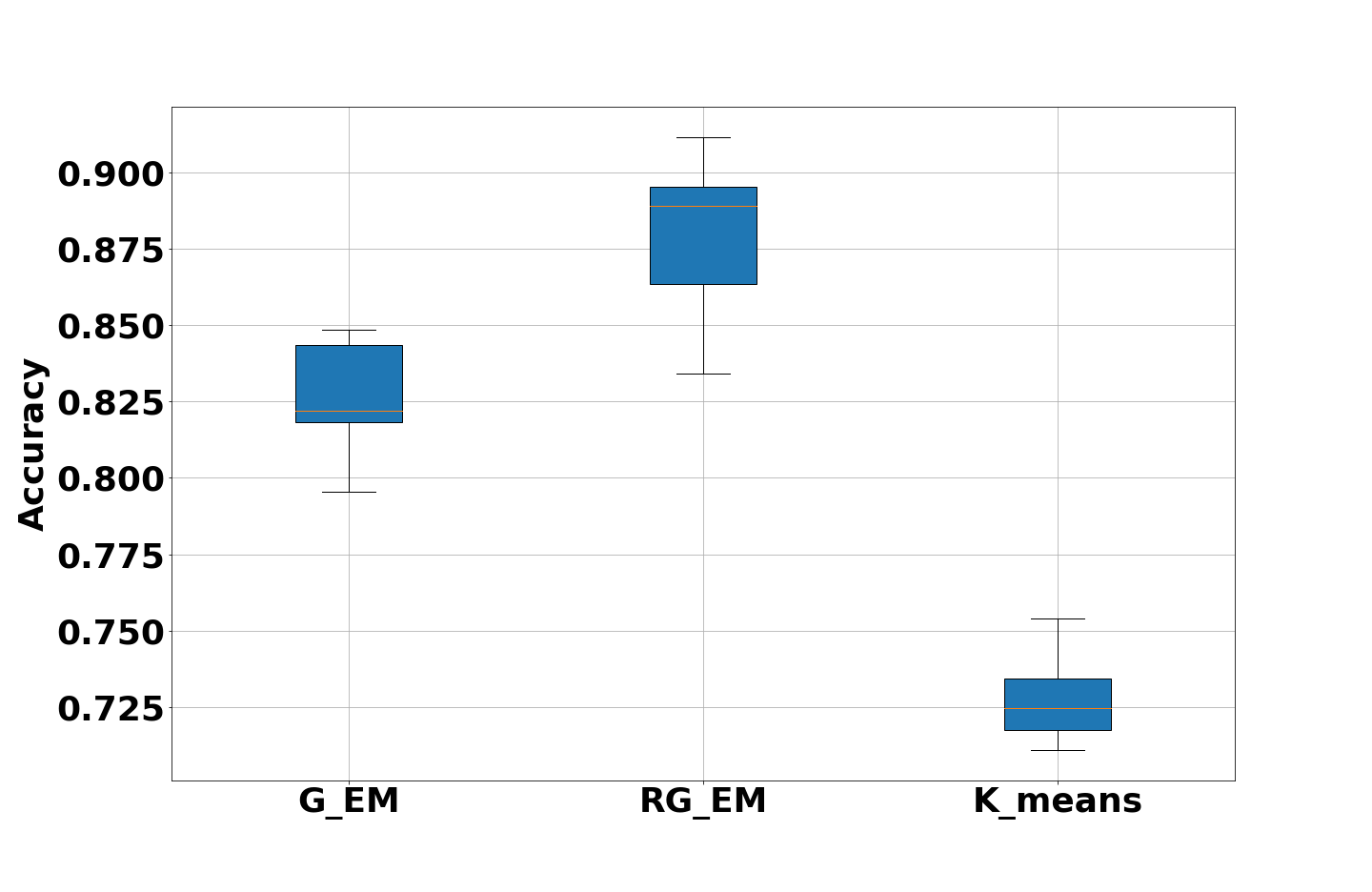}}
\caption{Pr\'ecision m\'ediane}
\label{fig:2} 
\end{figure}

On utilise 70\% des donn\'ees pour l'entraînement et 30\% pour l'\'evaluation des performances. Les r\'esultats sont moyenn\'es sur $100$ simulations et les datasets sont recompos\'es toutes les $10$ simulations. Utiliser une matrice cible circulaire n'est pas adapt\'e si certaines valeurs propres des matrices de covariance sont proche de 0. On effectue donc une analyse en composantes principales pour r\'eduire la dimension, on choisit la nouvelle dimension comme la plus petite permettant de conserver $95\%$ de l'information (variance). Cela correspond \`a $m=8$ pour Breast cancer et $m=26$ pour Ionosphere. Les nouvelles matrices étant proches de matrices diagonales, le choix de $\mathbf{T}_k = \hat \theta_k^0  \cdot \mathbf{I}_m$ semble pertinent. On obtient les r\'esultats pr\'esent\'es sur la figure \ref{fig:2}. 

\begin{figure}[!ht]
\centering
\subfigure[ionosphere\label{fig:3a}]{\includegraphics[scale=0.13]{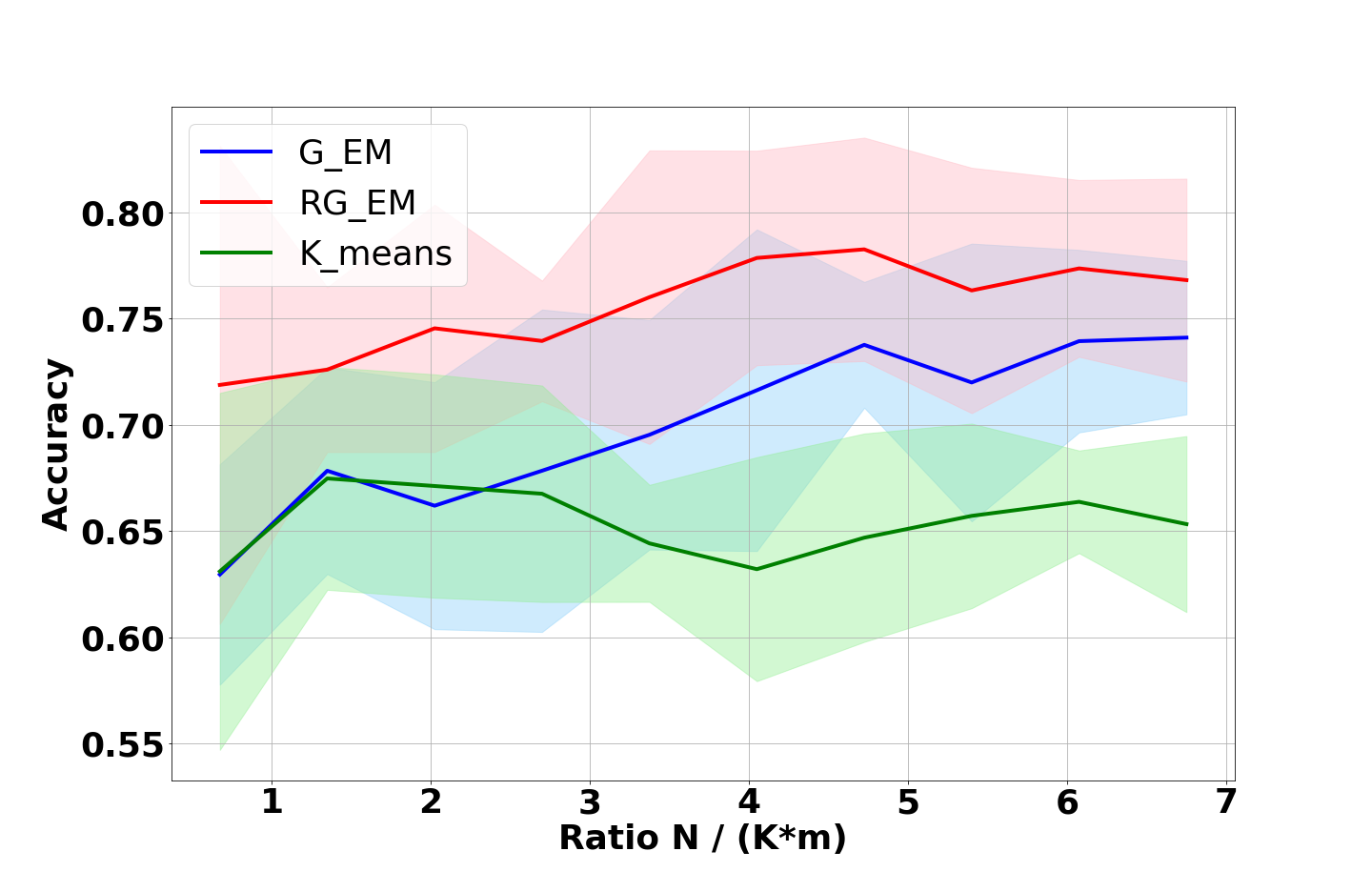}}
\subfigure[Breast cancer\label{fig:3b}]{\includegraphics[scale=0.13]{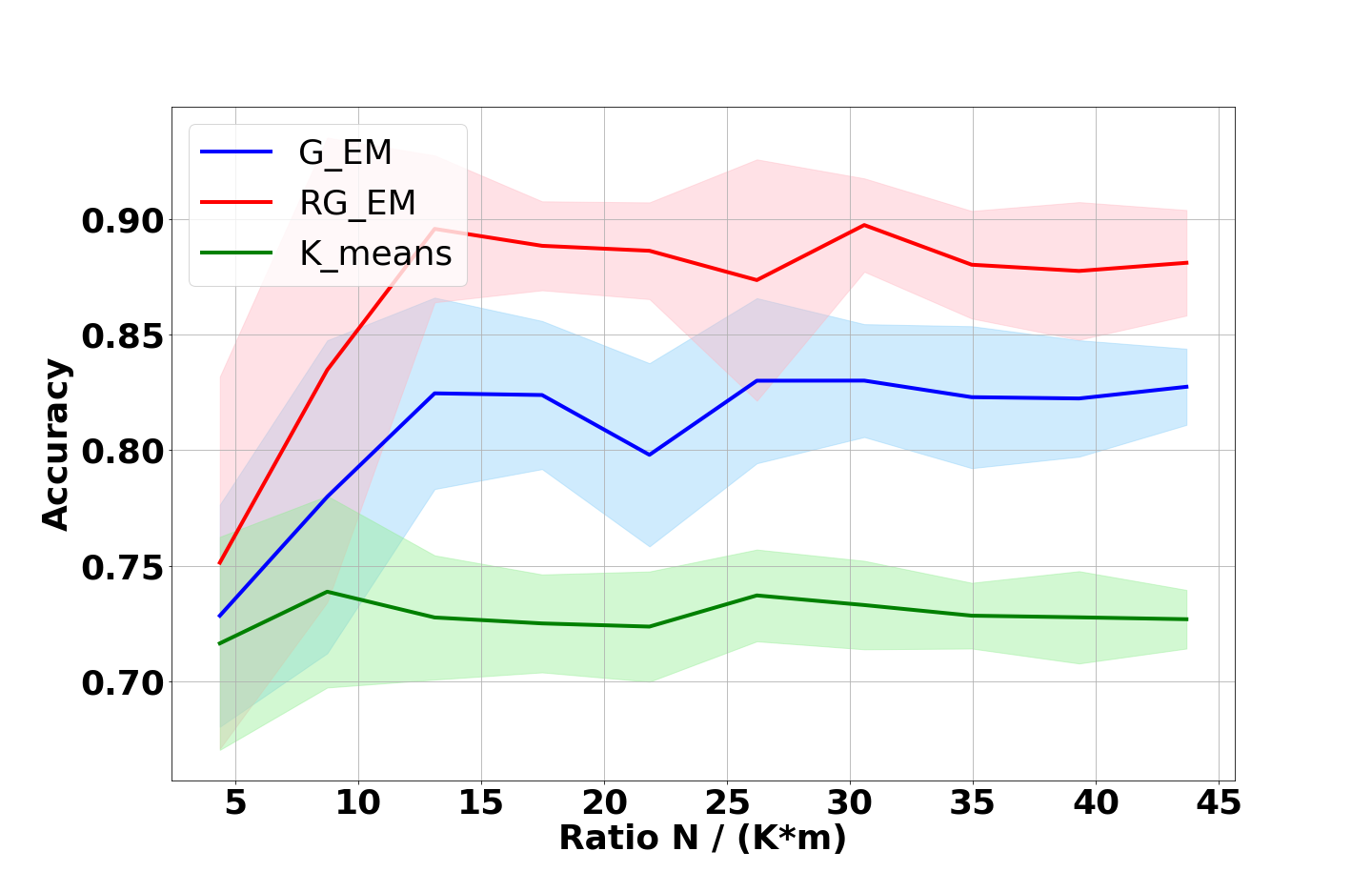}}
\caption{Evolution de la pr\'ecision en fonction du rapport $\frac{n}{m}$}
\label{fig:3} 
\end{figure}

Sur les deux jeux de donn\'ees, K-means obtient des performances sensiblement inf\'erieures aux m\'ethodes EM, avec un \'ecart d'environ $10\%$ de pr\'ecision. La version r\'egularis\'ee de l'EM conduit, sur les deux jeux de donn\'ees, à de meilleurs r\'esultats que l'algorithme GMM classique, la r\'eduction de dimension ayant rendu pertinent l'utilisation d'une matrice cible proportionnelle \`a l'identit\'e. On peut maintenant s'int\'eresser \`a l'\'evolution des performances de chaque m\'ethode lorsque le rapport $\frac{n}{m}$ devient de plus en plus faible. Pour observer cela, on supprime progressivement des donn\'ees du dataset d'entraînement pour r\'eduire sa taille de 100\% \`a 10\%, ce qui fait diminuer le rapport $\frac{n}{m}$. Les r\'esultats sont pr\'esent\'es sur la figure \ref{fig:3}.

Sur les deux jeux de donn\'ees, K-means n'est pas tr\`es impact\'e par la diminution du nombre de donn\'ees. En effet, la suppression des donn\'ees ne change pas la structure g\'eom\'etrique des clusters, et K-means construit une fronti\`ere similaire avec peu de donn\'ees. A l'inverse, les estimateurs des algorithmes EM sont impact\'es par la baisse du nombre de donn\'ees, ce qui provoque une diminution des performances. Sur le dataset breast cancer wisconsin, les deux m\'ethodes conservent des performances similaires jusqu'\`a ce que le nombre de donn\'ees soit r\'eduit de 80\%. La performance chûte alors rapidement pour rejoindre celle des autres m\'ethodes. Sur le dataset ionosphere, les deux algorithmes EM baissent progressivement, mais encore une fois, la version r\'egularis\'ee chute moins vite et conserve de meilleures performances.

\section{Conclusion}
\label{sec:conclusions}

Nous avons pr\'esent\'e dans cet article une version r\'egularis\'ee de l'algorithme EM-GMM qui surpasse les m\'ethodes classiques de clustering dans les r\'egimes o\`u le nombre de donn\'ees est faible par rapport à la dimension. Dans cette nouvelle approche, l'estimation de la matrice de covariance est r\'egularis\'ee avec un terme de p\'enalisation qui oriente l'estimation vers une matrice cible.  Les coefficients de r\'egularisation $\eta_k$ optimaux sont s\'electionn\'es gr\^ace \`a un algorithme de validation crois\'ee et r\'eguli\`erement mis \`a jour au cours des it\'erations. Les performances obtenues avec ce nouvel algorithme sont meilleures que celles obtenues avec des algorithmes classiques. De plus, la m\'ethode proposée, qui peut \^etre vue comme une am\'elioration de l'EM classique, est relativement stable en fonction du rapport $m/n$. Les perspectives de ces travaux vont se focaliser sur l'apprentissage des matrices cibles, ainsi que sur la version totalement non-supervisée de RG-EM.


\begin{thebibliography}{99}

\bibitem{Dempster2022Maximum}
A. P. Dempster and N. M. Laird and D. B. Rubin
\emph{Maximum Likelihood from Incomplete Data via the EM Algorithm}.
Journal of the Royal Statistical Society, 1977

\bibitem{Guorong2001EM}
Guorong Xuan and Wei Zhang and Peiqi Chai
\emph{EM algorithms of gaussian mixture model and hidden Markov model}.
Proceedings 2001 International Conference on Image Processing

\bibitem{Ingrassia2012Studies}
Ingrassia, Salvatore and Minotti, Simona C and Incarbone, Giuseppe
\emph{An EM algorithm for the student-t cluster-weighted modeling}.
Challenges at the Interface of Data Analysis, Computer Science, and Optimization, 2012

\bibitem{roizman2019flexible}
Roizman, Violeta and Jonckheere, Matthieu and Pascal, Fr{\'e}d{\'e}ric
\emph{A flexible EM-like clustering algorithm for noisy data}.
arXiv preprint arXiv:1907.01660, 2019

\bibitem{houdouin2022robust}
Houdouin, Pierre and Wang, Andrew and Jonckheere, Matthieu and Pascal
\emph{Robust classification with flexible discriminant analysis in heterogeneous data}.
ICASSP 2022

\bibitem{Teimour2021EM}
Mahdi Teimouri
\emph{EM algorithm for mixture of skew-normal distributions fitted to grouped data}.
Journal of Applied Statistics, 2021

\bibitem{Ying2014Regularized}
Ying Sun and Prabhu Babu and Daniel P. Palomar
\emph{Regularized Tyler{\textquotesingle}s Scatter Estimator: Existence, Uniqueness, and Algorithms}.
IEEE Transactions on Signal Processing, 2014

\bibitem{pascal2014generalized}
Pascal, Fr{\'e}d{\'e}ric and Chitour, Yacine and Quek, Yihui
\emph{Generalized robust shrinkage estimator and its application to STAP detection problem}.
IEEE Transactions on Signal Processing, 2014



\bibitem{yi2020shrinking}
Yi, Mengxi and Tyler, David E
\emph{Shrinking the Covariance Matrix using Convex Penalties on the Matrix-Log Transformation}.
Journal of Computational and Graphical Statistics, 2020

\bibitem{houdouin2022regularized}
Pierre Houdouin and Esa Ollila and Frederic Pascal
\emph{Regularized EM algorithm}.
https://arxiv.org/abs/2303.14989, 2022

\bibitem{UCI}
Dua Dheeru and Graff Casey.
\emph{{UCI} Machine Learning Repository}.
University of California, Irvine, School of Information and Computer Sciences, 2017.

\end{thebibliography}
\end{document}